\documentclass[journal]{IEEEtran}
\usepackage[ruled,vlined,linesnumbered]{algorithm2e}
\usepackage{amsmath}
\usepackage{graphicx}
\usepackage[noend]{algpseudocode}
\usepackage{color,soul}
\usepackage{subfigure}
\ifCLASSINFOpdf
\else
\fi
\begin{document}

%
\title{Understanding Human Innate Immune System Dependencies using Graph Neural Networks}

%
%
%

\author{Shagufta~Henna,~\IEEEmembership{Senior Member,~IEEE}
\thanks{Shagufta Henna is with the Computing Department, Letterkenny Institute of Technology, Letterkenny, Co. Donegal, Ireland.
e-mail: shaguftahenna@gmail.com

}
\thanks{Submitted under Fast Track: COVID-19 Focused Papers \newline
Manuscript received XX,XXXX; revised XX, XX.}}

%
%

\markboth{ }%
{Shell \MakeLowercase{\textit{et al.}}: Bare Demo of IEEEtran.cls for IEEE Journals}
%



\maketitle

\begin{abstract}

Since the rapid outbreak of Covid-19 and with no approved vaccines to date, profound research interest has emerged to understand the innate immune response to viruses. This understanding can help to inhibit virus replication, prolong adaptive immune response, accelerated virus clearance, and tissue recovery,  a key milestone to propose a vaccine to combat coronaviruses (CoVs), e.g., Covid-19. 
Although an innate immune system triggers inflammatory responses against CoVs upon recognition of viruses, however, a vaccine is the ultimate protection against CoV spread. The development of this vaccine is time-consuming and requires a deep understanding of the innate immune response system.  In this work, we propose a graph neural network-based model that exploits the interactions between pattern recognition receptors (PRRs), i.e., the human immune response system. These interactions can help to recognize pathogen‐associated molecular patterns (PAMPs) to predict the activation requirements of each PRR. The immune response information of each PRR is derived from combining its historical PAMPs activation coupled with the modeled effect on the same from PRRs in its neighborhood. On one hand, this work can help to understand how long Covid-19 can confer immunity where a strong immune response means people already been infected can safely return to work. On the other hand, this GNN-based understanding can also abode well for vaccine development efforts. Our proposal has been evaluated using CoVs immune response dataset, with results showing an average IFNs activation prediction accuracy of $90\%$, compared to $85\%$ using feed-forward neural networks. 
 
 \end{abstract}


%
\IEEEpeerreviewmaketitle

\section{Introduction}

\IEEEPARstart{C}ovid-19 patients show unique clinical/para-clinical features including fever, cough, shortness of breath, and chest abnormalities. The features associated with chest abnormalities can be detected by medical chest imaging including computed tomography (CT) or X-ray imaging techniques \cite{huang20}. These features, however, do not distinguish Covid-19 from the pneumonia \cite{ming20}. Early Covid-19 diagnosis is a real concern to facilitate the timely isolation of a suspected patient due to the unavailability of the Covid-19 vaccine.

Reverse transcription-polymerase chain reaction (RT-PCR)determines the volume of specific ribonucleic acids by interacting with ribonucleic (RNA) and deoxyribonucleic acids (DNA)\cite{pcr2020}. The test can detect severe acute respiratory syndrome coronavirus 2 (SARS-CoV-2) strain for Covid-19 detection. One of the major limitations of this test is negative at an initial stage which is detected as a positive by a CT scan \cite{Tali2020}. Several existing works have recommended CTS scans and X-rays as a better choice due to the limited availability of RT-PCR 
\cite{Tali2020,Narin2020,Maghdid2020,Bukhari2020}.
 
Another popular method to screen Covid-19 is real-time reverse transcription-polymerase chain reaction (rRT-PCR) \cite{Loeffelholz2020}. Although, rTR-PCR is capable to provide results within a few hours, however, its sensitivity is not high with values ranging from 37\% to 71\% \cite{Ali20}. Low sensitivity can result in a substantially high number of false-negative results at an early stage of infection. For such false-negative cases, recent works suggest chest radiology as a potential tool to detect Covid-19 \cite{Li2020}. Recently, the Fleischner Society has issued a consensus statement on the suitability of CT scans at an early stage in different clinical settings  \cite{Rubin2020}. Owing to the non-invasiveness and high sensitivity of CT scan-based Covid-19 detection, it is recommended for early Covid-19 detection and thus isolation. Radiologic diagnostic support is not available 24 hours and may have restrictions concerning location \cite{MossaBasha2020}. Further CT scan cannot differentiate the features of Covid-19 from the pneumonia features, thus causing uncertain predictions by radiologists. 

A rapid rise in Artificial Intelligence (AI) has necessitated the need for automatic Covid-19 techniques in medical imaging. Deep learning has revolutionized a rapid and early Covid-19 based on accurate analysis of chest CT images in the early stage as compared to radiologic diagnostic support. Recently, several works based on deep learning have been proposed for Covid-19 detection. In recent work in \cite{wang2020}, Wang and Wong proposed a deep learning model for Covid-19 called as COVID-Net. The proposed model achieved an accuracy of 83\% for Covid-19 prediction. A work in \cite{Rajpurkar2020} proposed a ChexNet model which has demonstrated outstanding results as compared to a radio-logistic diagnosis to detect pneumonia. 

A reliable deep learning screening method for Covid-19 detection based on CXR images is proposed in \cite{Zhang2020}. The work consists of backbone, classification, and detection components. The classification model is based on convolutional neural networks (CNNs) with the “sigmoid” activation function. The anomaly module calculates anomaly scores which detects anomaly images for Covid-19 to reduce the false-positive rate. The proposal demonstrates a sensitivity of 96\% with an accuracy of 71\%. Until recently, CNNs have been a popular choice for pneumonia and other chest diagnoses including Covid-19.  Alternative simpler solutions to CNN, e.g., ResNet-50 \cite{Gao2019} can process the images faster using numerous hidden layers. 

Scarselli et al. \cite{Scarsell09} introduced Graph neural networks (GNN) which is widely used for predictive tasks including node classification and link prediction. GNN can learn and exploit dependencies in sparse and relational structures in data \cite{Luca2019}. In GNN, a node can use recursive neighborhood aggregation to determine a new state and feature vector.

Research as have we discussed so far, continues to rely on deep learning-based Covid-19 detection as a potential research avenue.  Most of the existing deep learning approaches are based on convolutional neural networks including ResNet, DenseNet, VCGNet which can exploit a large number of hidden layers with extensive hyperparameter tuning and long training time. Although CNN-based models are preferred deep learning models due to the processing of large data sets, however, max pooling layer can only transfer information from one layer to the next layer.  It can result in loss of small details in the data and may not capture the data which cannot be transferred from one layer to another layer. Another limitation of CNN and other deep learning models is that it cannot capture the critical relationships which exist in data. In the context of Covid-19, there are dependencies between the innate human immune system and Covid-19 features which we can exploit to predict the Covid-19 detection without relying on X-ray images and CT scans. 

Unlike existing works, in this work, we model dependencies/relationships between PAMPs associated with PPRs using GNN with a particular focus on human innate immune system. 
Based on this model, we propose Covid-19 detection and learning mechanisms which can predict   Covid-19 with high sensitivity and accuracy.  Our proposed Covid-19 prediction Algorithm $Covid19_{gnn}$ can learn useful relationships between the PRRs of innate human immune system by using Algorithm $Covid19_{learn}$. The experimental evaluations and results based on PRRDB2.0 taken as a positive dataset and  Swiss-Prot as negative data set show that the proposed Algorithm $GNN_{covid}$ can predict PRRs activation with high sensitivity and accuracy compared with the PRRs activation traditional feed-forward neural network (FNN). Our evaluations reveal that the Covid-19-GNN IFNs activation can take into account useful dependencies between PRRs. Our results reasoned that the proposed $GNN_{covid}$ can extract useful insights into PRRs interactions that can be useful for Covid-19 detection with high accuracy even at an early stage.

The remainder of this paper is organized as follows. Section \ref{pftrs} details the immune response features. Section \ref{model} presents the proposed Covid-19 immune response system model based on GNN. Section \ref{gnnsp} describes the proposed Covid-19-GNN Algorithm with learning illustrated in Section \ref{learn}. Section \ref{eval} presents the performance evaluations of the proposed Covid-19-GNN activation prediction method. The conclusions and future works are given in Section \ref{conclude}.

%
%
%
%


\section{Immune Response Features:pathogen‐associated molecular patterns (PAMPs)} \label{pftrs}
The primary features with respect to human immune system are PAMPs. These PAMPs are recognized by different pattern recognition receptors (PRRs). Various PAMPs features associated with different PRRs are discussed below.

\subsection{Toll‐like receptors PAMPS}
Toll‐like receptors (TLRs) can recognize PAMPs denoted as $PAMPs_{tlrs}$ including lipids, proteins, lipoproteins, and nucleic acids.The recognition of $PAMPs_{tlrs}$ occurs in cell membranes, endosomes, lysosomes, endocytolysosomes, and other locations in cells \cite{akira2006}. Different TLRs can induce different biological responses via subsequent activation of varied adapter proteins, such as MyD88, TIRAP, TRIP, and TRAM. All these adapter proteins share  the Toll receptor (TIR) structure. MyD88 is the first identified TIR family member which acts as an adapter protein by almost all TLRs except TLR3 \cite{kawai2010}.

\subsection{RIG‐I‐like receptors PAMPs}
RIG‐I‐like receptors (RLRs) can recognize PAMPs features based on nucleic acids denoted as $PAMPs_{rig}$. $PAMPs_{rig}$ can result from different infections including Influenza A virus (IAV), Measles virus (MV), and Hepatitis C virus (HCV). These $PAMPs_{rig}$ can also include viral nucleocapsid proteins containing triphosphate and double basic acid RNA at the 5`‐end \cite{Goubau2014}. These features are used to identify  RNAs of picornaviruses, including poliovirus (PV) and Encephalomyocarditis virus (EMCV). These RNAs are primarily characterized by long double‐stranded RNA more than 1 kbp. 

\subsection{Nucleotide‐binding and oligomerization domain‐like receptors ((NLRs) PAMPS }
NLRs recognize PAMPs features $PAMPs_{nlrs}$ based on conserved Nucleotide‐binding and oligomerization domain (NOD) structure. These features can be based on various proteins/complexes called the inflammasome, reproduction, and regulatory NLRs. The inflammasome consists of at least eight NLR proteins, including NLRP1, NLRP3, NLRP6, NLRC4, NLRC5W, and AY2 
\cite{Davisbk11,Hornung09,VanGorp14}.

\subsection{C‐type lectin‐like receptor (CLRs) PAMPs}
CLRs recognize PAMPs features, denoted as $PAMPs_{clrs}$. CLRs are activated directly through macrophage‐induced Mincle and CLEC4E, and Dectin‐2 CLEC6A) receptors. The indirect activation of CLRs is triggered by the  HAM‐like motifs in the intracellular tail of the receptor, e.g.,
Dectin‐1 (CLEC7A) and DNGR‐1 (CLEC9A)
\cite{VanGorp14}. These activations result in acidified spleen tyrosine kinases which triggers formation of  CARD9, B‐cell lymphoid tissue 10 (BcL10), and Maltlcomplex formation. The signalling pathways can also include  SyK and JNK acidified apoptosis‐related protein granule, e.g., ASC 
\cite{hara2013}. These pathways in turn activate downstream $PAMPs$, i.e., NF‐kB and MAPKs. These $PAMPs$ can trigger various cellular responses, e.g., phagocytosis, maturation, and chemotaxis of cells \cite{hara2013}.

\subsection{Cytoplasmic DNA receptor PAMPs}

CLRs Cytoplasmic DNA receptor (CDR) can recognizes DNA CpG islands, denoted as $PAMPs_{cdr}$ \cite{Hemmi2000}. Examples of CDRs include AIM2‐like receptors (ALRs), DNA‐dependent activator of IFN‐regulatory factor (DAI), 
leucine‐rich repeat flightless‐interacting protein 1 (LRRFIP1), DExD/H‐box RNA helicase (DDX), Meiotic recombinant protein 11 Homolog A (MRE11), RNA polymerase III (Pol III), DNA dependent protein kinase (DNA‐PK), DNA repair‐related proteins Rad50 and Sry‐related HMG box 2 (Sox2) \cite{Chiu2009,Sun2013}. DAI recognizes PAMPs based on Z‐type DNA and B‐type DNA \cite{GallegoMarin2018,Jensen2012}. These DNAs depend on the length of DNA. AIM2-related PAMPs recognizes  double‐stranded DNA. IFI16 and cGAS receptors can recognize  cytosolic DNA recognition and are capable of  type I interferon \cite{Jensen2012}.

\subsection{Type I interferons}
Pathogen‐associated molecular patterns recognize the viral nucleic acid, activate IRF3 and IRF7 and promotes type I interferons (IFNs). IFNs trigger the JAK‐STAT signal pathway, thereby promoting 
IFN‐stimulated genes (ISGs) 
\cite{MaDY2015,Nelemans2019}. IFNs are antiviral molecueles which contribute a major role to immunomodulatory. Specifically, antigens resultant from these IFNS restricts infected target and T/B cells and any blockage to IFNs can effect the survival of the virus 
\cite{Ivashkiv2014,Cao2019}.

PRRs consists of three types including membrane,secretory and cytoplasmic \cite{Gordon2002}. The membrane PRRs include TLR2, TLR4, mannose receptor (MR) and scavenger receptor (SR). On the other hand, the secretory PRRs consists of mannose‐binding lectin (MBL) and C‐reactive protein (CRP). TLR3, TLR7/8 and NLRs form cytoplasmic PRRs. Among all these types, PRRs including TLRs, RLRs and NLRs result in IFN production.
SARS‐CoV and other coronaviruses are pathogenic and are sensitive to IFN-a/b. The N-protein of SARS-CoV is classified as an immune-escape-protein and is antagonist against host interferon response 
\cite{Bromberg2007,pan2011}.

A well-reported EV71 infection  down-regulates JAK1, p-JAK1 and p-TYK2. This down-regulation results in blockage of JAK-STAT signaling pathway thereby inhibits IFNs production. Reduced function of IFNs results in high EV71 replication in host cells 
\cite{liuy2014}. Similarly, Ebola virus (EBOV) inhibits IFNs production  by promoting cytokine signal inhibitory factor‐1 (SOCS1) which also blocks JAK-STAT signaling pathway 
\cite{Okumura2010}. 
Further, influenza A is also capable to inhibit IFN-I production by activating SOCS3 
\cite{Pauli2008}.

\subsection{Dendritic cells}

Dendritic cells (DCs)  stimulate the  the activation of T‐lymphocytes and B‐lymphocytes and play a vital role in innate and adaptive immunity. Mature DCs can activate T cells, thereby directly affecting the adaptive immune responses. DC precursor cells are differentiated based on inducers including  GM‐CSF, IL‐4, and TNF‐$\alpha$ if not transfected with HIV‐1 Nef protein. The viral antigens capability of DCs is limited by the HIV-1 which reduces the major histocompatibility antigen I (MHC I) on DC 's surface. The functionality of DCs is interfered by the viral infections which helps the viruses to evade from the adaptive immune response of the host \cite{Kaewraemruaen2019}. 

\subsection{Defensins}

Defensins consists of antibiotic peptide molecules which are critical for the host 's innate defense system. It eliminates bacteria including viruses, fungi, and tumor cells. Defensins are found in neutrophils which consists of small molecular cationic polypeptides. Defensin HNP‐1 inactivates viruses including HSV-1, HSV-2, cytomegalovirus (CMV), VSV, and IAV 
\cite{Zapata2016}. There are some studies which reveal that the human neutrophil defensin (HNP1‐3) are not able to inhibit or kill SARS‐CoV 
\cite{Zhux11}.


\section{Covid-19 Immune Response System Model} \label{model}

\subsection{Graph Neural Network Model}
We formulate the Covid-19 detection based on the PRR activation as a neighbor-dependency problem. The proposed solution which we call Covid-19-GNN uses Graph Neural Network (GNN)-based \cite{Scarsell09} approach to model the dependencies between PRRs. These PRRs dependencies can contribute towards the Covid-19 prediction. Given a human innate immune system with various PRRs which can recognize different PAMPs, the use of GNN is motivated by the fact that an immunity-response against Covid-19 results in inter-PRRs dependencies contributing to the impact of Covid-19 on the health of a particular individual.

A trained Covid-19-GNN can predict the intensity of Covid-19 based on the PAMPs features of a PRR and the PAMPs features of neighboring PRRs.PRRs are assumed to have PAMPs activation dependencies in the host innate immune system. The Covid-19-GNN consists of two feed-forward neural networks (FNNs). A PRR uses the first FNN of Covid-19-GNN to compute its next activation state denoted as $a_{n}$ based on the PAMPs features of its neighboring PRRs $n^{*}$. The second FNN predicts the activation of PRR based on the $a_{n}$ and its historical PAMP features. Covid-19-GNN model for Covid-19 intensity prediction is given below in form of Equation \ref{eq1} and Equation \ref{eq2}.

\begin{equation} \label{eq1}
a_{n}=\sum_{i \in n} h_{w} (f_{n},f_{i}, a_{i}), \forall n 
\end{equation}

\begin{equation} \label{eq2}
C_{n}(t)= g_{w} (a_{n},f_{n}), \forall n 
\end{equation}
The equation \ref{eq1} represents that the activation state function of a PRR depends on the PAMPs features $f_{i}$ and its activation state, i.e., $a_{i}$. In the equation, $h_{w}$ represents a parametric function that depends on its PAMPs features, PAMPs features of its neighboring PRRs, and the activation state of neighboring PRRs. The $g_{w}$ function in Equation \ref{eq2} denotes the activation prediction of PRR based on the activation state $a_{n}$  from equation \ref{eq1} and its historical PAMPs.The Covid-19-GNN consists of both $h_{w}$ and $g_{w}$ functions for each of the PRRs which converge exponentially fast. The fast convergence process can yield stable PRR activation states and predicted PRR activation which in turn can be used to predict the intensity of Covid-19.

\section{Covid-19 Graph Neural Network Model (Covid-19-GNN)} \label{gnnsp}

Given a host innate immune system which can activate PRRs to identify PAMPs to detect viral infections, the Covid-19-GNN aims to predict Covid-19 intensity.

The immune response of a PRR can adversely affect the activation of neighboring PRRs which reflects critical relationships among PRRs. It is possible to model these immune response PRRs dependencies with a GNN which can assist a faster and accurate Covid-19 immune response prediction which can be used to treat the patient promptly and helps to prioritize critical cases. These inter-dependencies among PRRs reflect the activation dependencies of the immune response system.

\begin{figure}[ht] 
    \centering \includegraphics[width=.45\textwidth]{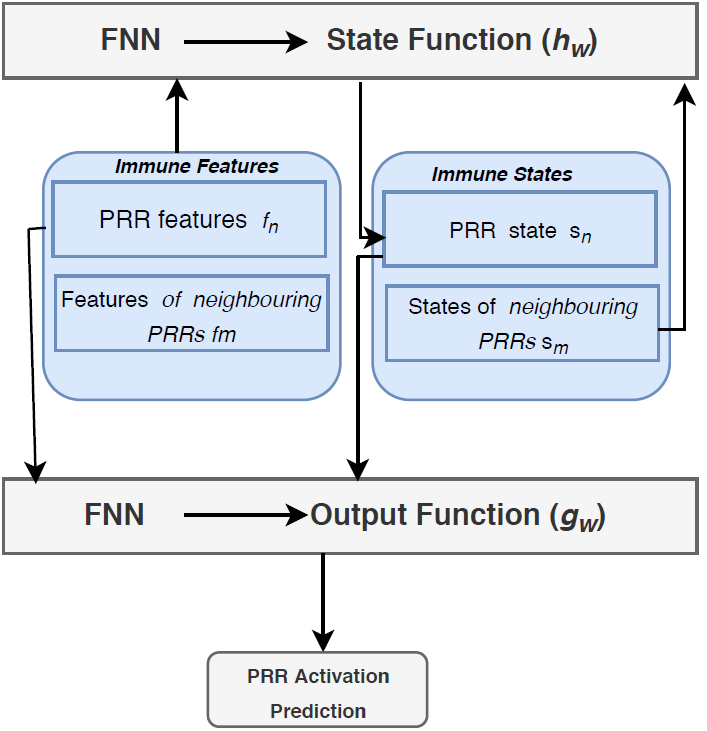}
\caption{GNN-based PRR activation.}\label{gconfi}
\end{figure}

GNN-based Covid-19 immune response is illustrated with the help of Equation \ref{eq1} and Equation\ref{eq2}. Figure \ref{coviddepex} shows an example of activation-related dependencies among PRRs. Figure \ref{gconfi} illustrates different modules of Covid-19-GNN. It consists of immune features, immune states, one FNN for $h_{w}$, and the other for $g_{w}$. The $h_{w}$ function provides the PAMPs features of a PRR and its state, whereas, the $g_{w}$ function predicts the activation state of PRR, e.g., IFNs based on its past features and state combined with the neighbouring PRR 's PAMPs features.

\subsection{Covid-19-GNN Features}
The Covid-19-GNN features include  PAMPs recognized by different PRRs. These PRRs include TLR, RLR, NLR, CLmin, cGAS, IFI16, STING, DAI, etc. Few of these have been discussed in detail in Section \ref{pftrs}. These PAMPs features are provided an input to predict the next activation state of PRR $a_{n}$ by using Equation \ref{eq1}. The output function in Equation \ref{eq2} takes the PAMPs of the current PRR $f_{n}$ and the $a_{n}$. For example, PAMPs recognized by TLRs $PAMPs_{tlrs}$ include lipids, lipoproteins, proteins, and nucleic acids of the bacteria and viruses.

 \subsection{Covid-19-GNN  States }

The PRRs are assumed to preserve the activation state $a_{n}$ derived by the parametric function $h_{w}$. This function takes the PAMPs features of a PRR and PAMPs of neighboring PRRs. $h_{w}$ shows that the activation state of a PRR depends on its PAMP features and PAMP features of its neighboring PRRs, thereby reflecting the activation dependencies of the immune response system.

The function  $h_{w}$in Equation \ref{eq1} computes the next activation state $a_{i+1}$ of PRR based on the current state $a_{i}$. It means, the current state $a_{n}(i)$ of PRR depends on the previous states $a_{m}(i-1)$ of its neighbour $m$ as given in Equation \ref{eq5}.

\begin{equation} \label{eq3}
a_{n}(i+1)=  \sum_{m \in n*} h_{w} (f_{n},f_{m},a_{m}(i))
\end{equation}

\begin{figure}[ht] 
    \centering
    \includegraphics[width=.35\textwidth]{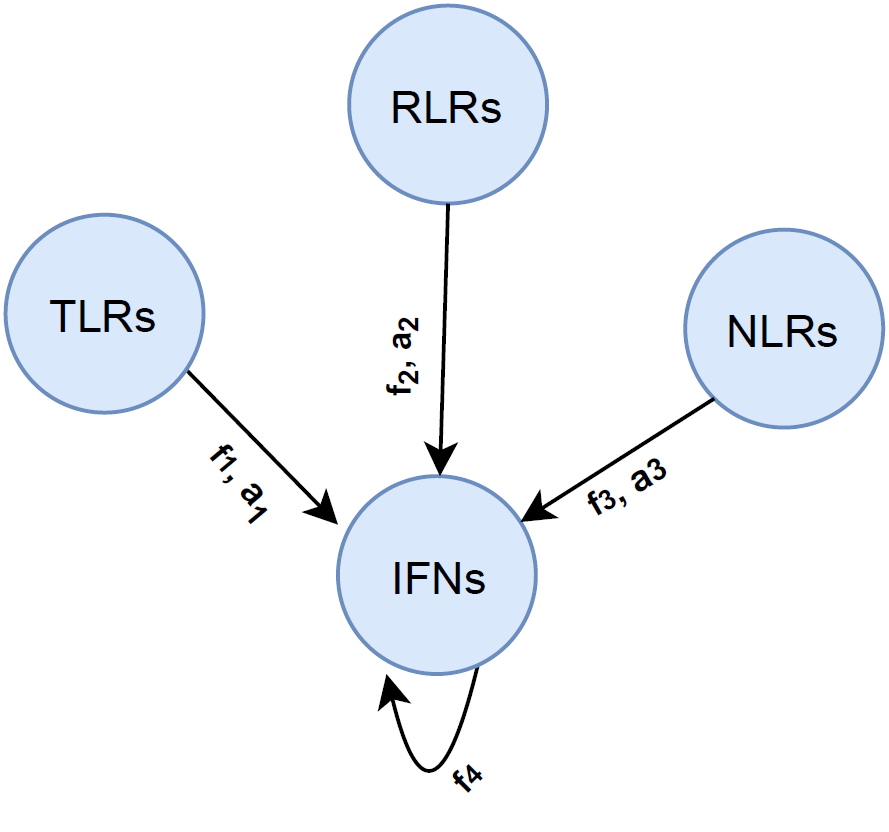} 
\caption{States \& Features from IFNs's neighbouring PRRs.}\label{coviddepex}
\end{figure}

Figure \ref{coviddepex} depicts the activation dependency of a PRR on the features of neighbouring PRRs. The figure shows that the state of the PRR IFNs depends on the activation states $a_{1}$, $a_{2}$, $a_{3}$ and features of TLRs, NLRS, and RLRs, i.e., $f_{1}$, $f_{2}$, $f_{3}$. The activation state of IFNs denoted as $a_{4}$ can be predicted based on Equation \ref{eq1}. The state $a_{4}$ at IFNs in figure \ref{coviddepex} is given as Equation \ref{eq4}.

\begin{equation} \label{eq4}
a_{4}= h_{w} (f_{4},f_{1}, a_{1})+  h_{w} (f_{4},f_{2}, a_{2})+  h_{w} (f_{4},f_{3}, a_{3})
\end{equation}

\subsection{Output Function }
Covid-19 intensity prediction by a PRR depends on the next activation state as computed by the $h_{w}$ function and a PRR \'s features. For the example in figure \ref{coviddepex}, the output function is based on a $g_{w}$ as given in Equation \ref{eq5}. The function $g_{w}$ is an FNN trained by the gradient descent method discussed later. The $g_{w}$ function of each PRR predicts its next activation state.

 \begin{equation} \label{eq5}
C_{4}(t)= g_{w} (a_{4},f_{4})
\end{equation}


\subsection{Covid-19-GNN Algorithm}
 The Covid-19-GNN model is illustrated with the help of Equation \ref{eq1} and Equation \ref{eq2}. It takes as an input PAMPs features of a PRR and outputs, for each PRR, an activation state. The model computes the activation state of each PRR. This computation is the result of an iterative process where each PRR has $h_{w}$ functions equivalent to the number of its neighboring PRRs in the innate immune system. The PRR activation contributes to determining the Covid-19 intensity prediction even at an early stage.

$h_{w}$ and $g_{w}$ functions are implemented as FNN. This is illustrated in Algorithm \ref{alg1}. The algorithm shows that the Covid-19-GNN-based prediction consists of: 
\begin{enumerate}

\item Observation of a PRR's PAMPs and its directly connected neighboring PRRs,
\item Computation of the next activation state of PRR based on the observed PAMPs features, and,
\item Use PAMPs observations from step 1 and activation state from step 2 to predict the IFNs activation which can help to determine the Covid-19 at an early stage.
\end{enumerate}

\begin{algorithm}
  \SetAlgoLined
 \KwData{$k=0$, state($k$)=0,number of iterations $M$}
 
 \KwResult{Predicted PRRs Activation}
    Observe PAMPs features for all PRRs and their neighbouring PRRs \\
          \Comment{Feature observation}  \\
              
                \While{$k < M$} {
     compute next activation state $a(k+1)$ using
  $a_{n}(k+1)=\sum_{m \in n} h_{w} (f_{n},f_{m}, s_{m}(k)), \forall n$\\
  $k\Leftarrow k+1$
 
    }
  \Comment {Predict PRRs activation} \\
   $C_{k}= g_{w} (s_{n}(k),f_{n}), \forall n$ \\
    
    $PRRs_{a} \Leftarrow C_{k}$\\
        return ($PRR_{a}$, $a_{n}$)
            
\caption{Covid-19-GNN} \label{alg1}

\end{algorithm}

\section{Covid-19-GNN learning} \label{learn}
 
Both the FNNs for the two functions $h_{w}$ and $g_{w}$ require training in order to make accurate predictions. This is accomplished with the help of input $Q$ and output  $P$ which is used to tune the weights of the FNNs. In our use-case of Covid-19, this requires a dataset for a human innate immune system including historic and current PAMPs features. The learning of FNNs minimizes the cost function as given in Equation \ref{eq6}.

\begin{equation} \label{eq6}
q_{x}=  \sum_{m \in M} \left(\frac{1}{2}(C_{n}-P_{n})^{2}+ \beta L(C_{n})\right)
\end{equation} 

Equation \ref{eq6} computes the error to minimize the cost function. The loss is added to the error function and can be scaled based on $\beta$. The value of $L$ penalizes the weights of the FNN whenever its output, i.e.,  $C_{n}$ exceeds the  $\mu$ \cite{Scarsell09}. Algorithm \ref{alg2} illustrates the learning algorithm based on gradient descent. The algorithm adjusts the weights for both the FNNs for each $h_{w}$ and $g_{w}$ in order to minimize the cost function $q_{w}$ given in Equation \ref{eq6}.

\begin{algorithm}
  \SetAlgoLined
 \KwData{$w=0$, $k$, Desired criteria $ACC$}
  \KwResult{Learned FNNs for $h_{w}$ and $g_{w}$ }
                 
                \While{$k < ACC$} {
                                     $a_{n}(k) \leftarrow $ Algorithm \ref{alg1} ($k$) \\
                     $C_{k} \leftarrow $ Algorithm \ref{alg1} ($k$) \\    
                   $\frac{\partial q_{x}}{\partial w}\leftarrow $ BPPgradient($q_{x}$, $w$) \\
                       $w(k+1) \leftarrow w(k)- \alpha\frac{\partial q_{x}}{\partial w} $ \\
                   $k\Leftarrow k+1$
                                    }
                  return learned $h_{w}$ and $g_{w}$
            
\caption{Covid-19-GNN learning} \label{alg2}

\end{algorithm}

The Algorithm \ref{alg2} tunes the weights $w$ of FNNs based on gradientdescent, each for $h_{w}$ and $g_{w}$ to minimize the cost given in Equation \ref{eq6}. The algorithm executes iteratively and updates the state and output as given in line 2 and line 3 by calling the Algorithm \ref{alg1}. Line 4 computes the gradient  $\frac{\partial q_{x}}{\partial w}$ of the cost function for $h_{w}$ and $g_{w}$ for weight $w$. Line 5 updates the weight as  $w(k+1) \leftarrow w(k)- \alpha\frac{\partial q_{x}}{\partial w} $ where $\alpha$ denotes the learning rate. Line 4 uses the backpropagation-through-time (BPTT)\cite{Scarsell09} to compute the gradient of the cost function for both the FNNs each for $h_{w}$ and $g_{w}$.

\section{Evaluations} \label{eval}

PRRs sequences are obtained from the database PRRDB2.0 \cite{prrdb2019} to constitute a positive dataset.PRRDB2.0 provides extensive information about unique PRRs and PAMPs. Each entry of the PRR in PRRDB2.0 has details of the PAMPs features. The negative dataset is based on random non-PRRs sequences from the Swiss-Prot. This is based on non-redundant positive and negative subsets according to the approach given in Bendtsen et al. \cite{Bendtsen2004} and Garg et al. \cite{Garg2008}. The strategy ensures that the subsets are dissimilar, i.e., less than 40\% similarity to aid with the unbiased training and testing of FNN and GNN. The performance of FNN and GNN is evaluated using a five-fold cross-validation technique. Training and test sets are based on positive and negative subsets. $4$ positive and $4$ negative subsets are combined for the training set, whereas, one positive and one negative subset forms the test set. This process is repeated $5$ times. In each fold, the positive and the corresponding negative subset acts as a test set exactly once.

\subsection{Performance metrics}
Covid-19-GNN evaluation is based on threshold-dependent and threshold-independent scoring parameters. Threshold-based scoring is used to calculate sensitivity, specificity, accuracy, and Mathhew 's correlation coefficient (MCC). Sensitivity is defined as correctly predicted positive PRRs divided by the total positive PRRs. On the other hand, specificity denotes the true negative rate of PRRs. Accuracy represents a model's ability to predict true positive PRRs. MCC calculates the correlation coefficient between actual and predicted PRRs. The threshold-independent performance metric area under the receiver operating characteristic curve (AUC) represents the plot between sensitivity and false-positive rate.

\subsection{Results and analysis}

\begin{figure}[ht] 
    \includegraphics[width=.5\textwidth]{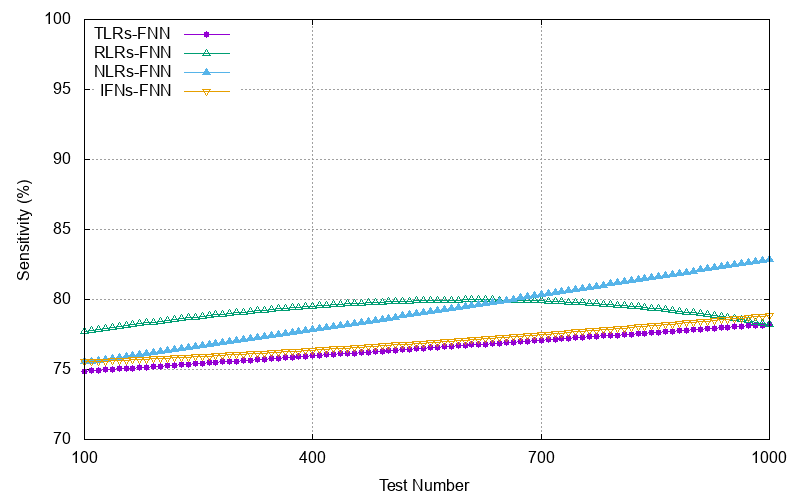}  
  
  \caption{Sensitivity of FNN.} \label{senseFNN} 
\end{figure}

\begin{figure}[ht] 
    \includegraphics[width=.5\textwidth]{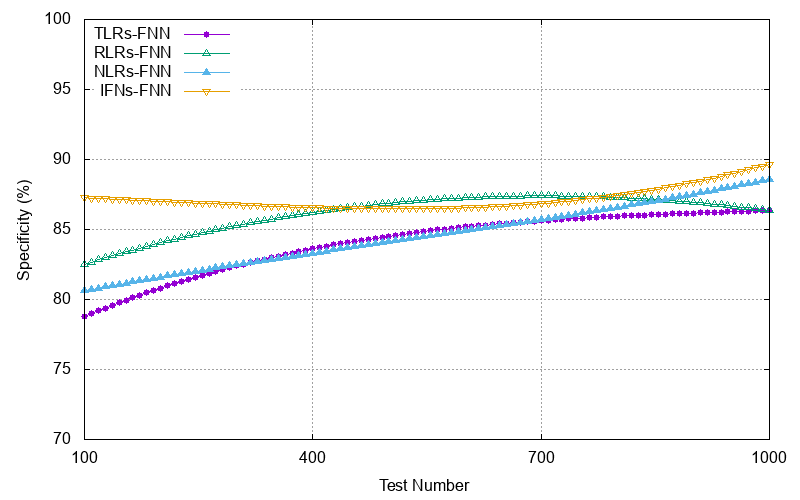}  
  
\caption{Specificity of FNN.} \label{speciFNN}
\end{figure}

\begin{figure}[ht] 
    \includegraphics[width=.5\textwidth]{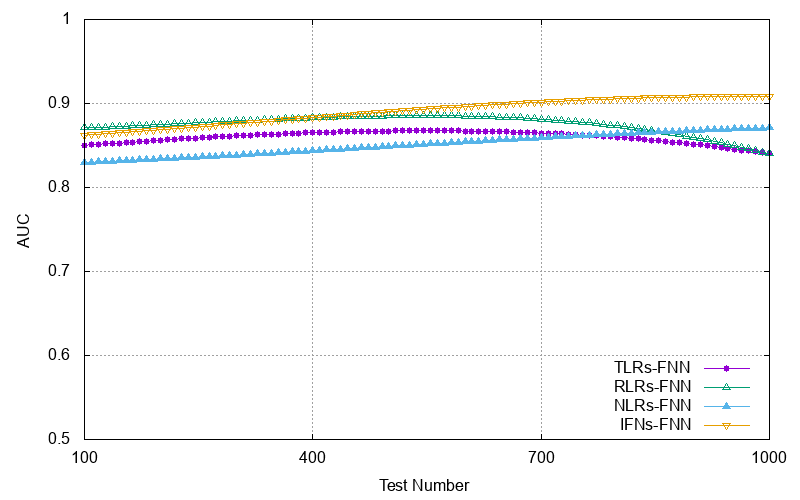}  
  
  \caption{AUC of FNN.} \label{aucFNN} 
\end{figure}

\begin{figure}[ht] 
    \includegraphics[width=.5\textwidth]{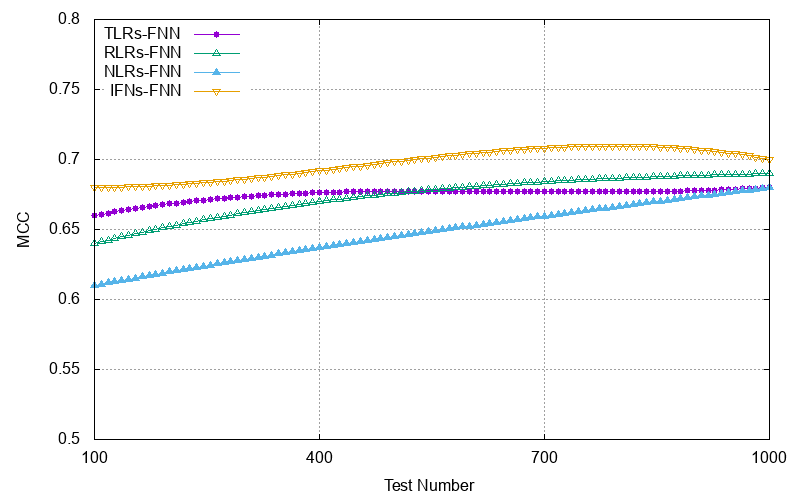}  
  
\caption{MCC of FNN.} \label{mccFNN}
\end{figure}

\begin{figure}[ht] 
    \includegraphics[width=.5\textwidth]{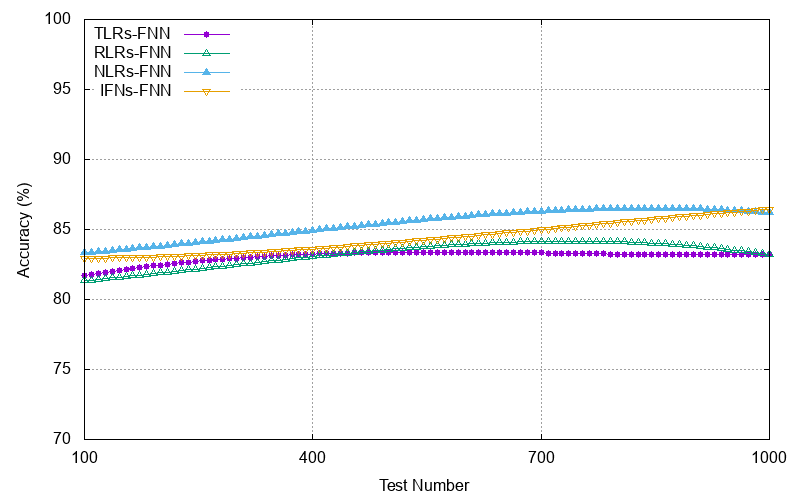}  
  
\caption{Accuracy of FNN.} \label{accuFNN}
\end{figure}

In the experiments, the trained Covid-19-GNN  system is tested to determine its sensitivity, specificity, prediction accuracy, MCC, and AUC over $1,000$ tests. The evaluation results for the PRR activation based on FNN are given in Figure \ref{senseFNN} to Figure \ref{accuFNN}.  Figure \ref{senseFNN} shows the sensitivity of the trained FNN-based Covid-19 to predict PRR activation. The trained traditional neural network achieves 78\% sensitivity and specificity of approximately 89\% for the activation of IFNs to predict innate immune response against Covid-19. The sensitivity for the NLRs activation is 83\% with a specificity of 89\%. TLRs's sensitivity associated with the TLRs using FNN varies 74\% to 76\% for all the test measurements with a specificity of a maximum 83\%. The Covid-19-GNN IFNs activation demonstrated an approximately 86\% sensitivity and 87\% specificity. The sensitivity of the TLRs activation based on FNN varies between 75\% to 77\%, with specificity between 76\% to 83\%. The Covid-19-GNN NLRs activation sensitivity is around  84\% for all the tests, with maximum specificity of 88\%. The maximum accuracy achieved by the FNN-based IFNs activation is approximately 85\%, with AUC 0.9\%, and MCC 0.74\%. On the other hand, the Covid-19-GNN IFNs activation accuracy is 90\% for all the tests, with AUC 0.9\%, and MCC 0.73\%.

\begin{figure}[ht] 
    \includegraphics[width=.5\textwidth]{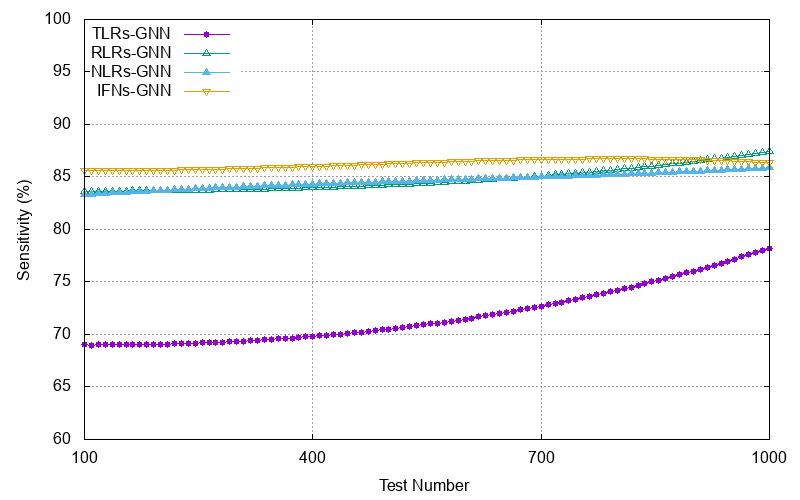}  
  
  \caption{Sensitivity of GNN.} \label{senseGNN} 
\end{figure}

\begin{figure}[ht] 
    \includegraphics[width=.5\textwidth]{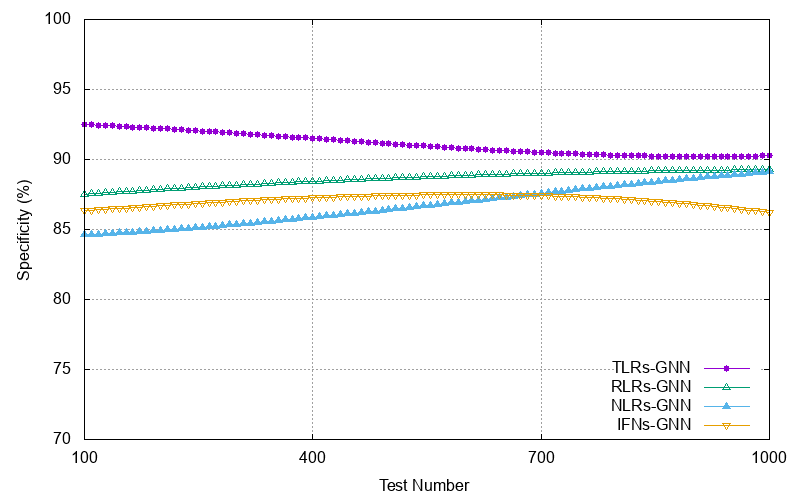}  
  
\caption{Specificity of GNN.} \label{speciGNN}
\end{figure}

\begin{figure}[ht] 
    \includegraphics[width=.5\textwidth]{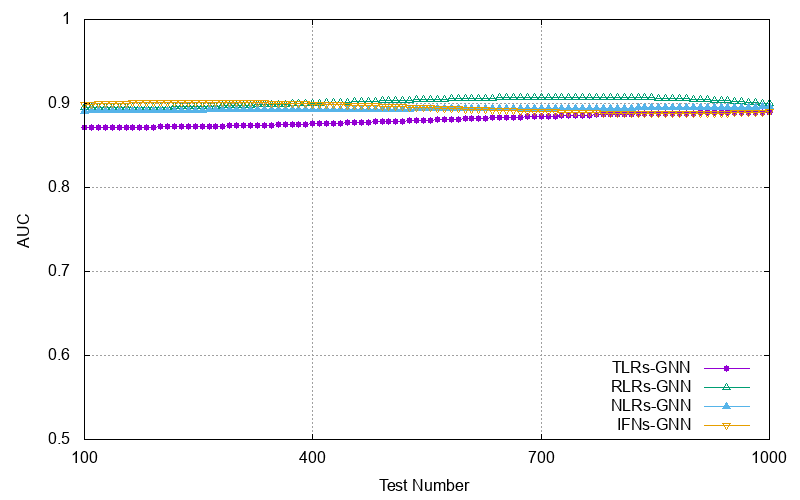}  
  
  \caption{AUC of GNN.} \label{aucGNN} 
\end{figure}

\begin{figure}[ht] 
    \includegraphics[width=.5\textwidth]{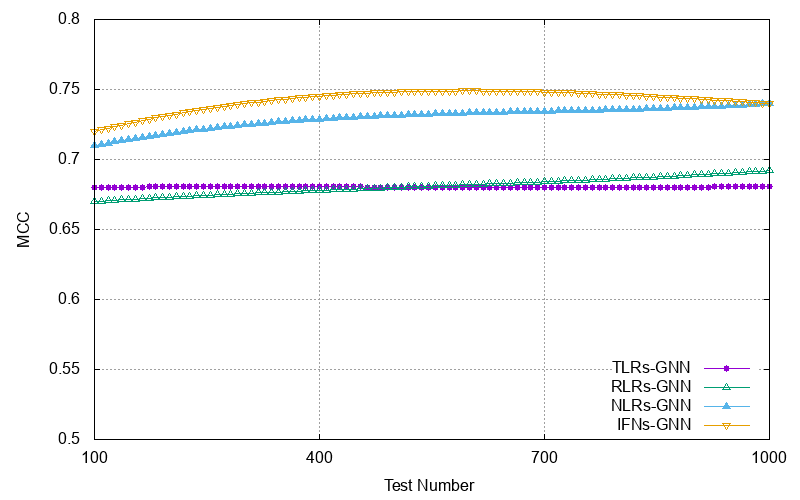}  
  
\caption{MCC of GNN.} \label{mccGNN}
\end{figure}

\begin{figure}[ht] 
    \includegraphics[width=.5\textwidth]{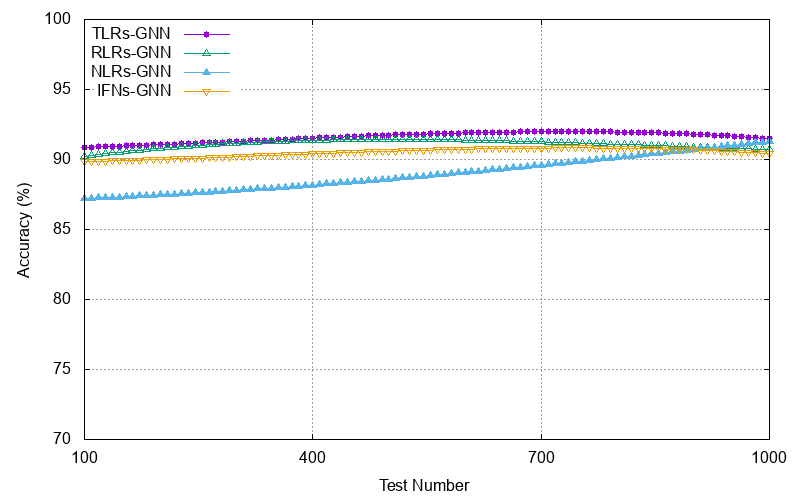}  
  
\caption{Accuracy of GNN.} \label{accuGNN}
\end{figure}

Since GNN model can specifically capture the features of the neighbouring PRRs to predict PRR activation, it performs better than FNN in terms of accuracy, specificity, and sensitivity. Given that a PRR 's activation can be predicted as a combined effect of PAMPs of its neighboring PRR and its own PAMP, the activation prediction accuracy and sensitivity is better than single FNN. We can observe from the Figure \ref{senseFNN}, Figure \ref{senseGNN}, Figure \ref{speciFNN}, and Figure \ref{speciGNN} that for the IFNs, sensitivity, and specificity are nearly equal which is an ideal result for the Covid-19-GNN PRR activation model. The AUC achieved by FNN-based activation for RLR, NLR, and TLR is approximately 0.85, while the Covid-19-GNN achieves an AUC of 0.9 for these PRRs as shown in Figure \ref{aucFNN} and Figure \ref{aucGNN}. This indicates the significance of our GCovid-19-GNN Algorithm for the action of PRR to predict the innate immune response against Covid-19. It can be observed from Figure \ref{mccFNN} and Figure \ref{mccGNN} that the MCC of FNN-based IFNs activation is  0.7. As compared to FNN-based IFNs activation, the GNN model shows higher performance with MCC nearly 0.75. The Covid-19-GNN NRLs activation demonstrates MCC of 74\% as compared to the FNN-based method with an MCC value of 0.74. Figure \ref{accuGNN} shows that the accuracy achieved by the Covid-19-GNN is significantly better than the FNN-based PRR activation, i.e., above 90\% for all the test cases.

\begin{figure}[ht] 
    \includegraphics[width=.5\textwidth]{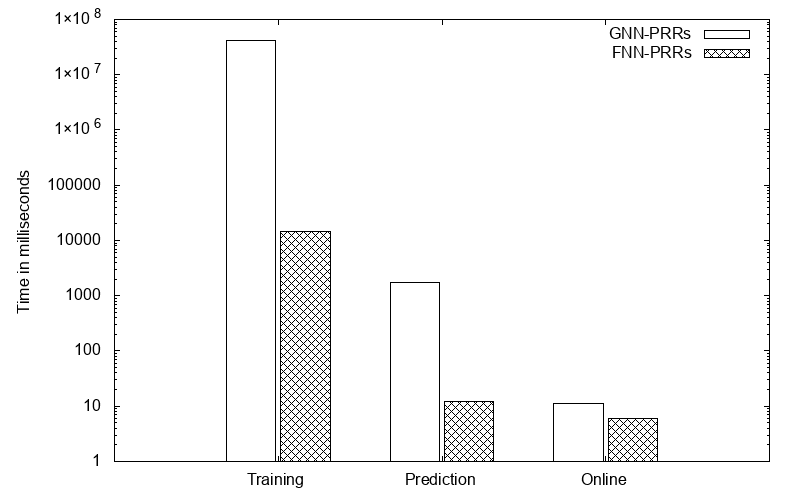}  
  
\caption{Training and prediction time of GNN.} \label{traintime}
\end{figure}

Figure \ref{traintime} compares the training time of Covid-19-GNN PRRs activation with the FNN-based method. The figure also shows the time taken by both the neural networks to make predictions. For both cases, GNN takes more time as compared to the FNNs. This additional time is attributed to the fact that GNN consists of multiple FNNs. It is evident from Figure \ref{traintime} that the GNN training time takes about 41,000 seconds. The GNN training process is offline which does not affect the online performance of the system to make predictions. Once trained, the weights of the FNNs are saved and loaded to make predictions. There is not a significant time difference for online predictions, e.g., for the time-critical systems predictions are kept running. These online predictions take about 11ms which is comparatively less than the offline training and prediction time.

From the above results, it is evident that the Covid-19-GNN PRRs activation outperforms the FNN-based activation method. This benefit is due to the topology-awareness of the GNN method. Another characteristic of GNN is its multiple FNNs with a higher number of layers, each with a higher number of neurons.

\section{Related Work}

The last few months have witnessed several Covid-19 investigations relevant to its detection and spread \cite{Bullock20}. The Covid-19 detection has been addressed based on different deep learning approaches using medical imaging including CT scans and chest X-ray. Other works discuss the spread of Covid-19 with a focus on the number of confirmed cases, recoveries, and deaths. 
Recent years have seen a rise in deep learning-based solutions to disease diagnosis, primarily based on x-ray images. In a similar effort, a deep neural network model called ChexNet for pneumonia detection based on chest x-ray images is proposed in \cite{Rajpurkar17}. The model demonstrated exceptional performance results in terms of accuracy. Following this work, another model called ChestNet  \cite{wang18}, a deep learning solution to predict thorax disease based on chest x-ray images.

In a recent work on Covid-19 diagnosis, authors have evaluated various convolutional neural networks (CNN) coupled with a pre-trained ResNet 50 model with 98\% accuracy \cite{Narin2020}. The evaluation study classifies healthy and Covid-19 infected patients. The work reports Covid-19 diagnosis with an accuracy of 97\% using InceptionV3 and 87\% based on Inception-ResNetV2. The study, however, does not consider the discrimination between pneumonia conditions from Covid-19.
In another work \cite{Apostolopoulos20}, authors presented a CNN-based solution to Covid-19 diagnosis from chest x-ray images. The proposal suggested a classification accuracy of approximately 97\% with MobileNet.

Wang et al. \cite{wang2020} proposed COVID-Net, a deep learning model for Covid-19 detection. The model demonstrated an accuracy of 83.5\% for Covid-19 classification based on healthy, bacterial-infection, and viral-infection classes. In another work in \cite{Hemdan2020}, authors evaluated different deep learning models for Covid-19 detection using chest x-ray images. The authors also proposed a COVIDX-Net model based on $7$ CNN models. In a similar work on Covid-19 detection \cite{Sethy2020}, authors have classified and evaluated the chest x-ray images as healthy or infected using various deep learning approaches, i.e., AlexNet, VGG16, GoogleNet, ResNet-101, and Inception-ResNet-v2, etc. The evaluations demonstrated an accuracy of approximately 95\% for RestNet50.

In \cite{Hassanien2020}, Hassanien et al. proposed a multi-level threshold-based SVM framework for classification. The proposed system detects Covid-19 using x-ray images. It relies on 40 x-ray images of 15 healthy and 25 infect cases. It has demonstrated an accuracy of 97\% with a sensitivity of 96\%. In a similar work in \cite{Gozes2020}, the authors demonstrated deep learning algorithms for Covid-19 detection based on CT scans of 157 patients. The system evaluated the deep learning-based detection algorithm using two subsystems. One subsystem considers a 3D analysis, whereas the second one performs 2D analysis of the CT scans using a Resnet-50-2 with an area under the curve of 99\%. The subsystem demonstrated a sensitivity of 98\% with a specificity of 92\%.

In recent work, Wang et al. \cite{wankang2020} proposed a deep learning solution to predict Covid-19 based on $195$ regions of interest (ROIs) of $395 \times 223$ to $636 \times 533$ pixels from the CT scans of $44$ Covid-19 positive patients and 258 ROIs from  50 Covid-19 negative patients. The internal validation of the proposed model has suggested an accuracy of $83\%$ with the specificity of $80.5\%$ and sensitivity of $84\%$. The proposed inception model demonstrates an accuracy of $73.1\%$ with a sensitivity of $74\%$ and a specificity of $67\%$. In \cite{Fu2020}, authors have proposed deep ResNet-50-based classification system for lung diseases including Covid-19 and pneumonia. The model is trained on more than 60K CT scans from 918 patients including Covid-19 and non-Covid-19. The proposed model has demonstrated an accuracy, specificity, and sensitivity of approximately 98\%. Another similar work in \cite{xu2019} reports image classification as Covid-19 and non-Covid-19 using RestNet-based methods. However, on the contrary to the work in \cite{Fu2020}, this work uses a Bayesian function to differentiate Covid-19 and non-Covid-19 images with an accuracy of 86.7\%.

\section{Conclusion and Future Work} \label{conclude}
 
In this work, we developed a graph neural network-based approach to predict the activation of IFNs in a human innate immune system to evaluate the intensity of Covid-19. The proposed model can model interactions between PRRs inherent to a human innate immune response system.  In comparison to FNN-based IFNs activation to predict Covid-19 severity, our model relies on dependencies between different PRRs to predict activation of IFNs. Our work can act as a milestone to investigate graph-based analytic to predict infectious disease through an understanding of the human innate immune response system. Future work aims to consider edge features of a graph in a GNN to weight the importance of each PRR interaction along with the PAMPs features. These interactions further can be modeled with self-supervised and supervised learning to predict PRRs activations for Covid-19 prediction at an initial stage.

\section{Data availability}

The datasets used for this study can be found at the
PRRpred webserver (https://webs.iiitd.edu.in/raghava/prrpred/
dataset.php).

\section*{Acknowledgement}
This work is supported by the Department of Computing, Letterkenny Institute of Technology, Letterkenny, Co. Donegal.

\ifCLASSOPTIONcaptionsoff
  \newpage
\fi



%

%

\begin{IEEEbiography}[{\includegraphics[width=1in,height=1.25in,clip,keepaspectratio]{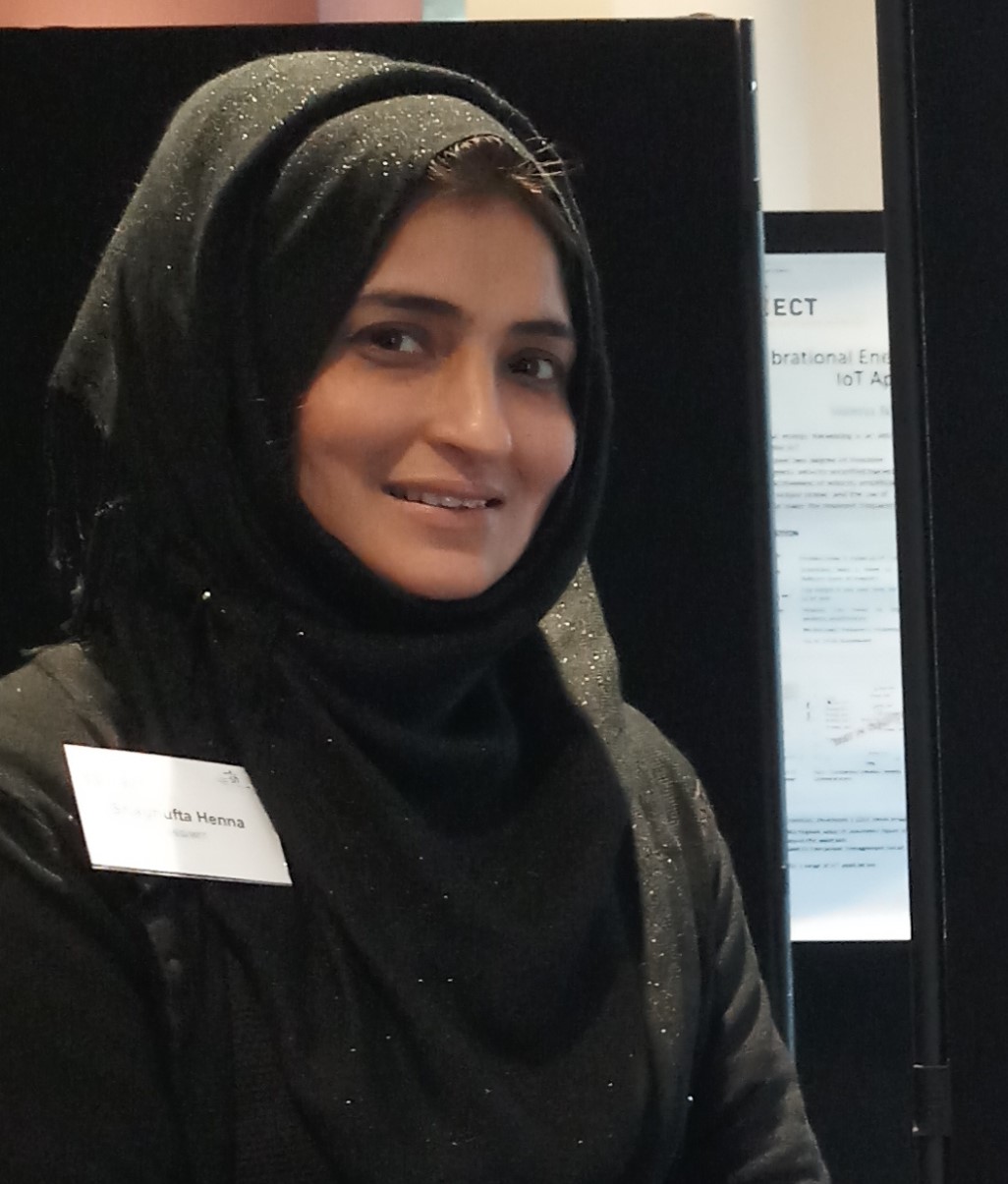}}]{Shagufta Henna} is an assistant lecturer with the Letterkenny Institute of Technology Co. Donegal, Ireland. She was a post-doctoral researcher with the telecommunication software and systems group, Waterford institute of technology, Waterford, Ireland from 2018 to 2019. She received her doctoral degree in Computer Science from the University of Leicester, UK in 2013. She is an Associate Editor for IEEE Access,EURASIP Journal on Wireless Communications and Networking, IEEE Future Directions, and Human-centric Computing and Information Sciences, Springer. Her current research
interests include big data analytics, distributed deep learning, and machine learning-driven network optimization.
\end{IEEEbiography}






\end{document}